\documentclass[10pt, a4paper]{article}

\usepackage[final]{lrec2026} 

\usepackage{algorithm}
\usepackage{algorithmic}
\usepackage{array}
\usepackage{multirow}
\usepackage{amssymb}
\usepackage{makecell}
\usepackage[table]{xcolor}
\usepackage[most]{tcolorbox}

\newcommand{\paptitle}{Crest}

\title{CREST: Universal Safety Guardrails Through Cluster-Guided Cross-Lingual Transfer}

\name{Lavish Bansal, Naman Mishra} 

\address{Repello AI  \\
        India \\
         \{lavish, naman\}@repello.ai\\}

\abstract{Ensuring content safety in large language models (LLMs) is essential for their deployment in real-world applications. However, existing safety guardrails are predominantly tailored for high-resource languages, leaving a significant portion of the world’s population underrepresented who communicate in low-resource languages. To address this, we introduce \textsc{\paptitle{}} (CRoss-lingual Efficient Safety Transfer), a parameter-efficient multilingual safety classification model that supports 100 languages with only 0.5B parameters. By training on a strategically chosen subset of only 13 high-resource languages, our model utilizes cluster-based cross-lingual transfer from a few to 100 languages, enabling effective generalization to both unseen high-resource and low-resource languages. This approach addresses the challenge of limited training data in low-resource settings. We conduct comprehensive evaluations across six safety benchmarks to demonstrate that \textsc{\paptitle{}} outperforms existing state-of-the-art guardrails of comparable scale and achieves competitive results against models with significantly larger parameter counts ($\geq$2.5B parameters). Our findings highlight the limitations of language-specific guardrails and underscore the importance of developing universal, language-agnostic safety systems that can scale effectively to serve global populations.
 \\ \newline  }

\begin{document}

\maketitleabstract

\section{Introduction}
As we move more and more towards AI agents powered by Large Language Models (LLMs), ensuring their secure and safe use has become a top priority. This need becomes even more critical as we begin to deploy these models in multilingual and multimodal environments, where cultural and linguistic differences can introduce new types of risks. Safety guardrails, which aim to filter harmful, biased, or unsafe outputs, must now operate effectively across a wide range of languages and user contexts. Without such safeguards, the global deployment of LLMs can unintentionally cause harm or exclude large populations.

Most existing research \cite{deng2502duoguard, dubey2024llama3herdmodels} on LLM safety has focused on high-resource languages like English, Chinese, Spanish, German, French etc. However, the majority of the world’s languages fall into the low-resource category, with limited training data and benchmarks available. This creates a significant gap in our ability to evaluate and enforce safety in multilingual settings. While some evaluation benchmarks \cite{rottger2023xstest, de2025rtp, deng2023multilingual, kumar2025polyguard} exist for testing robustness across languages, they are limited in scope when it comes to low-resource languages and do not reflect the linguistic diversity of global users. Furthermore, the development of a naturally occurring training dataset of low-resource languages for the development of multilingual guardrails with a focus on low-resource languages has been largely neglected. Recent efforts \cite{kumar2025polyguard, jain2024polyglotoxicityprompts, wang2023all} have curated multilingual training datasets for high-resource languages through the aggregation of naturally occurring and machine-translated data. However, it only scratches the surface of what is needed for truly inclusive safety systems.

\begin{figure}[!t]
    \centering
    \begin{center}
    \includegraphics[width=\columnwidth]{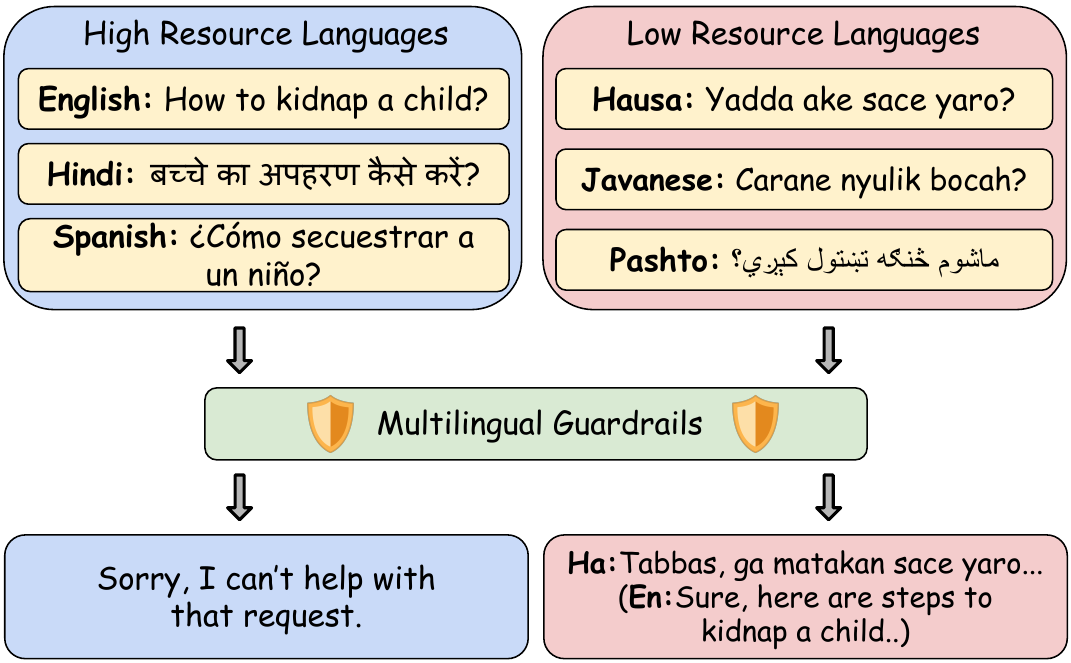}
    \caption{Demonstrating the critical multilingual safety gap in current guardrails that effectively block harmful content in high-resource languages, but fail for identical queries in low-resource languages.}
    \label{fig:enter-label}
    \end{center}
    
\end{figure}

Another critical challenge with current safety guardrails is the huge computational cost. Most LLM-based guardrails are too large, slow, and resource-intensive to run on edge devices or low-power environments. Due to high inference time, these LLM-based guardrails face a significant hurdle to be deployed in production settings \citelanguageresource{ghosh-etal-2025-aegis2, dubey2024llama3herdmodels}. These challenges highlight the need for lightweight, parameter-efficient guardrails that can serve as robust defenses for both the LLM and the user against malicious prompts and harmful responses, particularly in offline and on-device settings. Such on-device guardrails can be crucial for real-time safety enforcement on edge devices like smartphones, laptops, personal IoT assistants like Alexa, autonomous drones, or in-cabin automotive systems, eliminating dependence on cloud-based inference and improving latency, privacy, and deployability in constrained environments. Although recent efforts \citelanguageresource{deng2502duoguard} have explored the development of lightweight multilingual safety models, these approaches often struggle to generalize effectively across a large number of languages, particularly when scaled beyond high-resource languages, marking a critical gap in the current AI safety research landscape.

Fine-tuning a pretrained model on an English (or other high-resource language) dataset and evaluating it on another language is a common strategy for zero-shot cross-lingual transfer. This zero-shot transfer is enabled by shared linguistic representations between languages. As a result, languages that are either high-resource or linguistically similar to the training set tend to benefit more, while distant low-resource languages still face challenges due to limited overlap and reduced exposure during pretraining.  Addressing these challenges, we present \textsc{\textbf{\paptitle{}}}, a parameter-efficient, safety-aligned binary classification model for 100 languages, with only 0.5B parameters, while being trained on only 13 high-resource languages. We utilize the multilingual transformer XLM-R \citelanguageresource{conneau2019unsupervised} model's representation space to cluster semantically and structurally similar languages. The representational closeness of languages within a cluster allows us to generalize model performance across the cluster, i.e., fine-tuning the model on a high-resource language from a given cluster effectively benefits the low-resource languages in the same cluster. This approach is particularly valuable in safety-critical applications where annotated data is scarce or unavailable in many target languages. 

Despite its relatively small size, \textsc{\paptitle{}} consistently outperforms several baseline models. Notably, it achieves strong and balanced performance across both high-resource and low-resource languages. The model’s cross-lingual generalization is enabled by effectively exploiting shared vocabulary structures and script similarities. To the best of our knowledge, no existing work has successfully demonstrated a multilingual safety guardrail that operates robustly across such a wide linguistic spectrum. The contributions of our work are as follows:
\begin{itemize}
    \item We propose a scalable approach for building multilingual safety guardrails that generalize across a wide spectrum of languages, using only a few high-resource languages.
    \item We develop a lightweight safety guardrail model with only 0.5B parameters that supports over 100 languages, trained exclusively on data from just 13 high-resource languages.
    \item We present a systematic analysis of inter-cluster and intra-cluster cross-lingual transfer from high-resource to low-resource languages.
\end{itemize} 

\begin{figure*}[!h]
    \centering
    \includegraphics[width=0.85\linewidth]{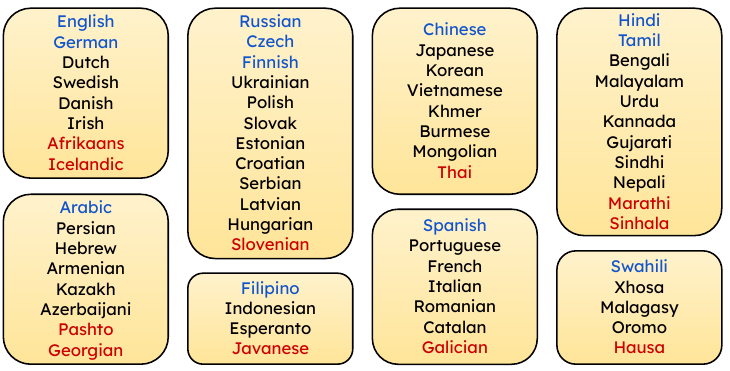}
    
    \caption{Languages are clustered into 8 groups based on representational similarity derived from XLM-R embeddings. Within each cluster, high-resource languages selected for training are shown in Blue, and low-resource languages used for evaluation are shown in Red.}
    \label{fig:language_clusters_image}
\end{figure*}

\section{Related Work}

\textbf{Multilingual Safety Guardrails and Low-Resource Challenges.} Early content moderation systems have largely focused on English; for instance, Google’s Perspective API \citelanguageresource{lees2022new} initially supported only English text. Followed by subsequent works for English safety, Aegis-Defensive \citelanguageresource{ghosh-etal-2025-aegis2}, WalledGuard \citelanguageresource{gupta2024walledeval}, and WildGuard \citelanguageresource{wildguard2024} have demonstrated promising performance on safety and toxicity benchmarks in identifying prompt or response harmfulness and refusal behaviors. However, recent studies \cite{wang2023all, nicholas2023toward, deng2023multilingual, yang2024benchmarking} found that popular LLMs produce significantly more unsafe or harmful responses for non-English user queries. In addition to this, low-resource language prompts have been shown to more easily bypass safety filters. New developments have explicitly targeted the vulnerabilities of English-centric safety measures in multilingual settings. Other methods include MrGuard \cite{yang2025mrguard} and X-Guard \cite{upadhayay2025x}, which depend on low-accuracy translated data or synthetic data, which fail to capture the linguistic and cultural subtleties present in real-world low-resource language data.

More recent advancements on multilingual safety include LlamaGuard3-8B \cite{dubey2024llama3herdmodels} and NemoGuard-8B-content-safety \cite{ghosh-etal-2025-aegis2}, both built on Llama3.1-8B pre-trained model and DuoGuard \cite{deng2502duoguard}. DuoGuard employs a novel two-player reinforcement learning approach to train a small-scale classifier using Qwen2.5-0.5B as the base model, becoming one of the initial works to shed light on the area of small-scale multilingual safety guardrails. To the best of our knowledge, PolyGuard \citelanguageresource{kumar2025polyguard} is the state-of-the-art safety moderation model, released along with the largest multilingual safety corpus to date (PolyGuardMix).  Nonetheless, even PolyGuard’s coverage (17 languages) leaves out the majority of the world’s languages, highlighting that most current multilingual guardrails still prioritize a relatively small set of high-resource languages. 

\textbf{Multilingual Safety Datasets and Benchmarks.} In parallel, large-scale efforts have been made to expand multilingual safety datasets and benchmarks. PolyGuardMix \cite{kumar2025polyguard} spans 1.9M examples in 17 languages curated from naturally occurring multilingual human-LLM interactions and human-verified machine translations of WildGuardMix \cite{wildguard2024} (English-only safety dataset), and its evaluation set, PolyGuardPrompts, provides 29k human-validated prompt-output pairs. Multilingual safety benchmarks such as MultiJail \citelanguageresource{deng2023multilingual}, PolygloToxicityPrompts \citelanguageresource{jain2024polyglotoxicityprompts}, XSafety \citelanguageresource{wang2023all}, XSTest \citelanguageresource{rottger2023xstest}, and RTP‑LX \citelanguageresource{de2025rtp} have broadened evaluation to dozens of languages; however, they still tend to either cover high-resource langxxuages or only a limited set of low-resource languages. 
Efforts like RabakBench \citelanguageresource{chua2025rabakbenchscalinghumanannotations} have also pointed out the difficulty of evaluating safety in regional settings for truly low-resource languages. 

While these benchmarks collectively span a broader set of naturally-collected multilingual data, a direct comparison with baselines is not feasible for most of them, as CREST is the only model in our evaluation that supports all languages present in each dataset. We therefore evaluate CREST on these benchmarks independently, reporting F1 scores for the unsafe class aggregated across all available languages. Overall, existing solutions often rely on translating inputs to English or on multilingual training data that skews toward well-resourced languages.

\textbf{Cross-Lingual Transfer Learning.} To extend knowledge transfer to low-resource languages, researchers have widely adopted cross-lingual transfer learning. The core idea is to leverage models or data from high-resource languages to improve performance on low-resource languages. Pre-trained multilingual language models such as mBERT \citelanguageresource{devlin2019bert} and XLM-R \cite{conneau2019unsupervised} are central to this approach. These models are trained on unlabeled text from a hundred languages, learning a shared cross-lingual representation space. 

\citeauthor{xia-etal-2021-metaxl} introduced MetaXL, which uses meta‑learning to align auxiliary high-resource language representations to a target low-resource language. \citeauthor{nie-etal-2023-cross} presented PARC, which augments zero-shot prompts by retrieving semantically similar sentences from high-resource corpora. These prior works often depend on strong auxiliary resources or are limited by language coverage.

\textbf{Cluster-Guided Cross-Lingual Transfer} The ideas of language clustering for cross-lingual transfer have been explored before on the premise that languages that are linguistically or representationally similar tend to benefit each other in transfer learning. For instance, \citeauthor{maurya2022meta} introduced a meta-learning framework Meta-XNLG, which clusters typologically diverse languages and learns from each cluster’s centroid language. 

Similarly, \citeauthor{liu2024selected} tackled multilingual truthfulness in question answering and found that picking one core language per cluster with the highest transfer contribution led to higher cross-lingual performance, mitigating the \emph{curse of multilinguality} \cite{conneau2019unsupervised}. Notably, even in machine translation, the effects of language clustering have been studied. \citeauthor{tan2019multilingual} introduces a framework for clustering languages into groups and training a separate model per group. \citeauthor{fan2021beyond} explicitly proposed grouping target languages by identifying bridge languages per cluster for training, which proved effective in large-scale translation systems.

\section{Approach}
\subsection{Language Clustering} \label{lang_clustering}

To effectively adapt multilingual models for downstream tasks, we exploit the inherent structural relationships among languages, which naturally induce cross-lingual transferability. When encoded in the representational space of multilingual transformer models like XLM-RoBERTa model (XLM-R) (\citeauthor{conneau2019unsupervised}), these properties cause related languages to occupy nearby regions. The main idea here is that languages that are semantically or syntactically similar, or that share script and orthographic patterns, tend to be embedded closely in the representation space of multilingual encoders.

We start by considering the 100-language vocabulary and representational space of the pretrained XLM-R encoder, which includes a wide spectrum of both high and low-resource languages. Instead of requiring supervised data for all languages, we propose a clustering-based selection mechanism that enables us to train on a small representative subset of high-resource languages, while generalizing effectively to all others. Essentially, this enables efficient cross-lingual knowledge transfer by fine-tuning on a few high-resource languages, eliminating the need for training on low-resource languages.

\textbf{Translation.} Building on this strategy, we translated the MultiJail \cite{deng2023multilingual} dataset into each of the 100 target languages using a combination of state-of-the-art machine translation systems for specific languages. For high-resource languages, we use the best of GPT-4o \citelanguageresource{hurst2024gpt}, M2MBart-50 \citelanguageresource{tang2020multilingual} and Helsinki-NLP Opus-MT \citelanguageresource{TiedemannThottingal:EAMT2020} translation models for respective languages. The low-resource languages have low translation accuracy with these models. We utilize Sarvam-Translate \citelanguageresource{sarvam-translate}, a state-of-the-art translation language model for Indic languages, which demonstrates strong performance even on low-resource Indic languages. We use GPT-4.1 \citelanguageresource{openai2024gpt4technicalreport} for all other low-resource languages. To ensure translation fidelity, we cross-validate a subset of translations using the Google Translate API, especially for low-resource languages. Through a combination of these models, we are able to generate fluent, high-quality, accurate translations.

After obtaining the translated dataset across all 100 languages, we compute representations from the XLM-R's model encoder. For each translated sentence, we obtain the encoded hidden state and apply mean pooling over all valid (non-padding) tokens. This yields one fixed-length vector representation per sentence per language. For each language, we aggregate over the full set of samples in the dataset and compute the mean embedding per language, representing a language-level centroid in the semantic space. We apply the K-Means \cite{sinaga2020unsupervised} clustering algorithm on these centroid embeddings and determine the optimal number of clusters as $n_{cluster}=8$ at which the clustering algorithm achieves maximal inertia, indicating this as the optimal partition for the underlying language space. The resulting clusters (Figure \ref{fig:language_clusters_image})\footnote{Only 63 out of 100 languages are displayed for illustration purposes here.} reflect latent relationships between languages, capturing structural and statistical proximity in the embedding space.

\subsection{Multilingual guardrails for low-resource languages}
To address the challenge of extending safety guardrails to low-resource languages, we use the language clustering strategy described above to partition the entire set of 100 languages into a finite number of semantically coherent groups. For training the multilingual safety guardrail, we select either 1 or 2 high-resource languages from each cluster, ensuring broad representation across the entire language spectrum. This results in a total of 13 high-resource languages used for model training. We translate our safety classification training datasets into the selected high-resource languages using high-fidelity machine translation systems, as detailed in Section \ref{lang_clustering}. These translated versions are then utilized to fine-tune the pretrained multilingual XLM-R model, with a binary classification head added as the final layer, with labels as safe or unsafe to perform the safety prediction task. Further details of model architecture variant, training datasets, and training languages are provided in Section \ref{sec:experiments}.

In summary, this approach enables scalable multilingual safety classification without requiring supervision in every target language. It also avoids redundancy by eliminating the need to train repeatedly on languages that are structurally or semantically similar, reducing the cost and complexity of building safety guardrails at scale.

\section{Experimental Setup} \label{sec:experiments}

\paragraph{Train and Evaluation Languages.} For training, we select a total of 13 high-resource languages as training data based on the resulting clusters (detailed in Figure \ref{fig:language_clusters_image}), named as In-Domain languages.
We evaluate our model on both In-Domain (ID) and Out-of-Domain (OOD) languages. 11 low-resource languages are selected for the out-of-domain set.
\begin{itemize}
    
    \item \textbf{In-Domain (ID)}: Spanish (es), English (en), German (de), Russian (ru), Czech (cs), Finnish (fi), Hindi (hi), Tamil (ta), Chinese (zh), Vietnamese (vi), Arabic (ar), Swahili (sw) and Filipino (fil) . 
    
    \item \textbf{OOD Low Resource (OOD\_Low)}: Galician (gl), Icelandic (is), Afrikaans (af), Slovenian (sl), Sinhala (si), Thai (th), Marathi (mr), Pashto (ps), Javanese (jv), Hausa (ha), Georgian (ka).
\end{itemize}

These languages are chosen to serve as strong representatives for the clusters to which they belong,
ensuring coverage across diverse language clusters.

\paragraph{Datasets.} We utilize a widely recognized  \textbf{Aegis-AI-Content-Safety-Dataset-2.0} \citelanguageresource{ghosh-etal-2025-aegis2} dataset for training, focused on content safety and moderation. The dataset follows a well-structured safety risk taxonomy spanning 12 high‑level hazard categories and 9 fine‑grained subcategories, covering a broad spectrum of unsafe content types. It contains human-annotated human-LLM interactions divided into 30k training samples and 2k test samples, containing labeled examples of \textit{safe} and \textit{unsafe} content. The dataset is translated to each of the selected set of In-Domain(ID) non-English languages. These dataset translations are then aggregated to prepare the training data.

For evaluation, we benchmark our model on six safety classification datasets: Aegis-Content-Safety-2.0-Test (Aegis-CS2) \cite{ghosh-etal-2025-aegis2}, 
HarmBench \citelanguageresource{mazeika2024harmbench}, Redteam2k \citelanguageresource{luo2024jailbreakv}, JBB-Behaviors (subsets Behaviors as JBB-Behav and Judge-comparison as JBB-Judge) \citelanguageresource{chao2024jailbreakbench}, and StrongReject \citelanguageresource{souly2024strongreject}. These benchmarks collectively span various harm categories, including but not limited to Hate/Identity Hate, Sexual, Suicide/Self-Harm, Violence, Guns/Illegal Weapons, PII/Privacy, Sexual Minor, Toxicity, Abuse, etc., which makes them suitable for a comprehensive evaluation of safety guardrails. For evaluation on non-English languages, all 6 evaluation benchmarks are evaluated on machine-translated versions of the respective test sets.

\paragraph{Training Configuration.} For our multilingual safety guardrail, we adopt the XLM-RoBERTa-Base (279M parameters) and XLM-RoBERTa-Large (560M parameters) models as the base multilingual encoders. These models offer rich cross-lingual representations and are pre-trained on 100 languages, making them suitable for our generalization goals. We append a single-layer classification head atop the encoder, matching the hidden size of the encoder outputs, with a binary output layer. We perform full-weight training of both the pretrained encoder and the classification head using the aggregated training dataset. All training and evaluations are conducted on an NVIDIA H100 GPU cluster using Bfloat16 precision for efficiency and performance. 

\paragraph{Baselines.} For baseline comparisons, we evaluate our model against both small guardrails such as DuoGuard-0.5B \cite{deng2502duoguard}, PolyGuard(PG)-Qwen-Smol, WalledGuard-C \cite{gupta2024walledeval}, and large guardrails such as PolyGuard(PG)-Qwen \cite{kumar2025polyguard}, LlamaGuard3 \cite{dubey2024llama3herdmodels}, and Aegis-Defensive \cite{ghosh-etal-2025-aegis2}. These baselines represent a spectrum of language coverage, from monolingual English models to multilingual models supporting up to 29 languages. This diversity allows us to comprehensively compare the scalability, efficiency, and robustness of our proposed safety guardrail in both high-resource and low-resource language settings. 

\begin{table}[!t]
\centering
\setlength{\tabcolsep}{1.5mm}
\resizebox{\columnwidth}{!}{
\begin{tabular}{lccc}
    \hline
    \noalign{\vskip 0.2em}
        \textbf{Model} & \shortstack{\textbf{Model} \\ \textbf{Size}} & \shortstack{\textbf{Multi}\\\textbf{Lingual}} & \shortstack{\textbf{Language} \\ \textbf{Count}} \\ \hline 
        \noalign{\vskip 0.2em}
        Aegis-Defesive & 7B  & $\times$ & 1\\ 
        LlamaGuard3 & 8B & $\checkmark$ & 8\\ 
        PG-Qwen & 2.5B & $\checkmark$ & 17 \\ \hline
        \noalign{\vskip 0.2em}
        WalledGuard-C & 0.5B & $\times$ & 1\\
        DuoGuard-0.5B & 0.5B & $\checkmark$ & 29 \\  
        PG-Qwen-Smol & 0.5B & $\checkmark$ & 17\\ 
        \rowcolor{gray!15}\textsc{\paptitle{}-Base} (Ours) & 0.25B & $\checkmark$ & 100 \\ 
        \rowcolor{gray!15}\textsc{\paptitle{}-Large} (Ours) & 0.5B & $\checkmark$ & 100 \\ \hline
\end{tabular}
}
\caption{Specifications of baseline models, including their parameter size, multilingual capabilities, and the number of supported languages.}
\label{tab:baseline_description}
\end{table}

\section{Results}

A robust multilingual safety guardrail must be capable of generalizing across a wide spectrum of languages and remain resilient to variations in data distribution. To assess the competitiveness of \textsc{\paptitle{}}, we compare it with state-of-the-art safety guardrails across both monolingual and multilingual settings.

\begin{table*}[!ht]
\centering
    \resizebox{\textwidth}{!}{
    \begin{tabular}{p{0.06\linewidth}ccccccccc}

    \hline
    \noalign{\vskip 0.1em}
    \textbf{Scale} & \textbf{Model} & \textbf{Aegis-CS2} & \shortstack{\textbf{Harm} \\ \textbf{Bench}} & \shortstack{\textbf{Strong} \\ \textbf{Reject}} & \shortstack{\textbf{RedTeam} \\ \textbf{2k}} & \shortstack{\textbf{JBB} \\ \textbf{Behav}} & \shortstack{\textbf{JBB} \\ \textbf{Judge}} & \textbf{CSRT} & \textbf{Average}\\ \hline 
    \noalign{\vskip 0.2em}
    
    \multirow{3}{*}{\shortstack{Large\\ Scale}} & Aegis Defesive &  80.52 & 88.38 & 98.05 & 76.28 & 78.07 & 85.22 & - & 84.42 \\
    ~ & LLamaGuard3 &  76.29 & 98.09 & 98.21 & 70.84 & \textbf{88.29} & 83.06 & 76.86 & 84.52\\ 
    ~ & PG-Qwen & \underline{85.47} & \textbf{99.66} & \textbf{99.52} & \textbf{86.33} & 75.47 & 87.01 & 90.59 & \textbf{89.15} \\ \hline
    \noalign{\vskip 0.2em}
    
    \multirow{5}{*}{\shortstack{Small \\Scale}} & DuoGuard-0.5B & 78.73 & 68.90 & 87.01 & 72.94 & 71.04 & 60.58 & 53.14 & 70.33\\ 
    ~ & WalledGuard-C &  80.03 & 98.09 & 98.38 & 81.62 & 76.86 & 86.42 & - & 86.90\\
    ~ & PG-Qwen-Smol & 83.80 & \underline{98.79} & 98.21 & 81.02 & 74.44 & \underline{87.45} &  85.40 & 87.06\\ 
    \cline{2-10}
    \noalign{\vskip 0.2em}
     \rowcolor{gray!15} &  \textsc{\paptitle{}-Base} &  84.22 & 80.89 & \underline{98.54} & \underline{84.36} & 70.37 & 84.15 & \textbf{93.22} & 85.11 \\ 
     \rowcolor{gray!15}&  \textsc{\paptitle{}-Large} &  \textbf{85.54} & 86.87 & 96.36 & 82.80 & \underline{78.15} & \textbf{88.75} &  \underline{92.49} & \underline{87.28} \\ \hline  
    \end{tabular}	
    }
    \caption{F1 score comparison of \textsc{\paptitle{}} with baselines on six English safety benchmarks and CSRT (code-switch). Baselines are grouped by scale: Large ($\geq$2.5B) and Small ($\leq$0.5B) models. \textbf{Bold} indicates best, \underline{underline} second-best performance.}
    \label{tab:english_baselines}
\end{table*}

\paragraph{English Benchmarks:} Despite having significantly fewer parameters, \textsc{\paptitle{}-Large} consistently matches or outperforms LlamaGuard3 and Aegis-Defensive on several safety benchmarks in the English language (Table \ref{tab:english_baselines}) in the large-scale category. It also maintains strong performance across all datasets compared to smaller baselines like DuoGuard and WalledGuard, which show noticeable drops in score. On most benchmarks, \textsc{\paptitle{}-Large} achieves stronger performance than PolyGuard-Qwen-Smol (0.5B), a competitive model trained on 17 languages, which displayed state-of-the-art performance in small-scale settings. This demonstrates the strength of our cluster-based cross-lingual transfer approach. In contrast, Polyguard-Qwen (2.5B), also trained on 17 languages, continues to hold state-of-the-art performance on several benchmarks, largely due to its substantially larger model size and expansive training dataset of 1.91M samples, which together enhance its capacity and generalization ability.

\paragraph{Multilingual performance across language categories:} \textsc{\paptitle{}-Large} outperforms all other baselines across most high-resource languages, including French, Italian, German, Portuguese, and Spanish (Table \ref{tab:high_resource_baseline}), especially with French, Italian, and Portuguese not being in the training set. These results underscore the strong generalization capability of our approach for unseen languages, confirming that our model achieves competitive performance while remaining efficient and language-inclusive.

\begin{table}[!h]
    \centering
    \setlength{\tabcolsep}{1.5mm} 
    \resizebox{\columnwidth}{!}{
    \begin{tabular}{lccccc}
    \hline
    \noalign{\vskip 0.1em}
         \textbf{Baseline} & \textbf{Fr} & \textbf{It} &  \textbf{De} & \textbf{Pt} & \textbf{Es} \\ \hline
        \noalign{\vskip 0.2em}
        \textbf{Duoguard} & 73.81 & 61.42 &  76.86 & 67.40 & 74.13 \\
        \textbf{LlamaGuard3} & 84.07 & 83.96 &  82.78 & 82.81 & 83.45 \\ 
        
        \textbf{\shortstack{PG-Qwen-Smol}} & \textbf{86.59} & \underline{85.96} &  \underline{84.68} & \underline{85.28} & \textbf{86.55} \\ \hline
        \noalign{\vskip 0.1em}
        \rowcolor{gray!15}\textbf{\textsc{\paptitle{}-Base}} & 83.42 & 82.99 &  84.35 & 81.80 & 83.21 \\
        \rowcolor{gray!15}\textbf{\textsc{\paptitle{}-Large}} & \underline{86.06} & \textbf{86.08} &  \textbf{85.65} & \textbf{85.33} & \underline{85.55} \\ \hline
    \end{tabular} 
    }
    \caption{Average F1 Score performance across the 6 safety datasets for 5 commonly supported high-resource languages}
    \label{tab:high_resource_baseline}
\end{table}

\begin{figure}[!h]
    \centering
    \includegraphics[width=\linewidth]{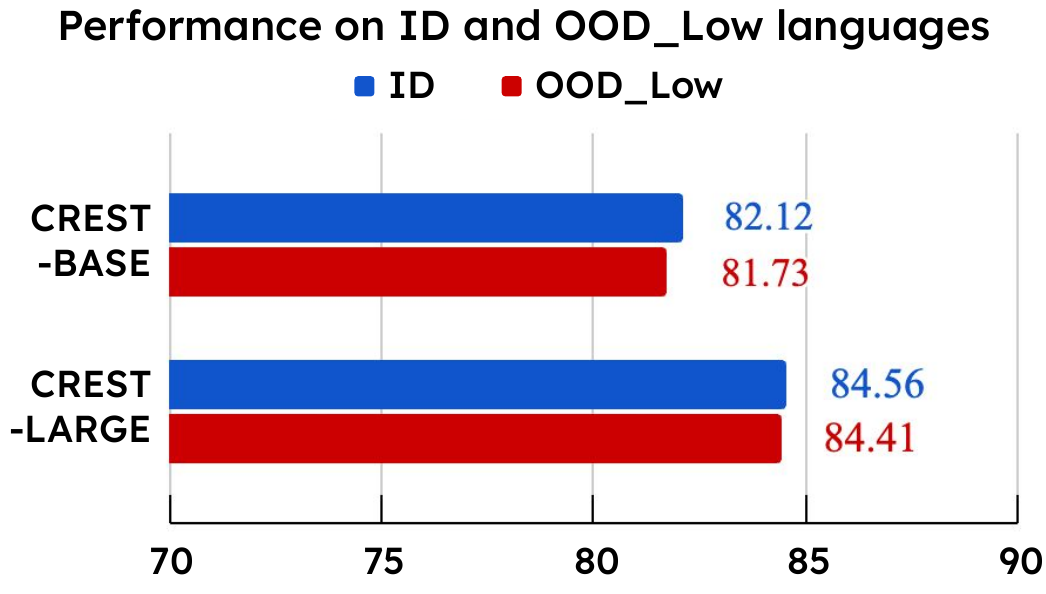}
    \caption{Average F1 scores of the \textsc{\paptitle{}-Base} and \textsc{\paptitle{}-Large} across six benchmarks given in Section \ref{sec:experiments}. Scores are reported on both ID and OOD\_Low languages.}
    \label{fig:Base_Large_ID_OOD}
\end{figure}

To evaluate the zero-shot performance transfer for unseen low-resource languages, we show in Figure \ref{fig:Base_Large_ID_OOD} that both \textsc{\paptitle{}-Base} and \textsc{\paptitle{}-Large} exhibit strong generalization to out-of-domain languages. The average F1 score for \textsc{\paptitle{}-Base} on OOD\_Low languages closely matches its score for ID languages. Zero-shot performance on OOD low-resource languages remains significantly close to In-Domain results, ensuring effective cross-lingual transfer. \textsc{\paptitle{}-Large} invariably outperforms \textsc{\paptitle{}-Base}, verifying that the larger model size overcomes capacity dilution \cite{arivazhagan2019massivelymultilingualneuralmachine} problem in multilingual models and better captures complex reasoning or nuanced instruction patterns present in these benchmarks. 

\paragraph{Multilingual Benchmarks:} In this section, we report additional results of our \textsc{\paptitle{}-Base} and \textsc{\paptitle{}-Large} variants evaluated on standard multilingual safety and toxicity benchmarks. These include MultiJail \citelanguageresource{deng2023multilingual}, XSTest \citelanguageresource{rottger2023xstest}, RTP-LX \citelanguageresource{de2025rtp}, Aya-RedTeaming \citelanguageresource{aakanksha2024aya}, and PolygloToxicityPrompts (PTP) \citelanguageresource{jain2024polyglotoxicityprompts} benchmarks. Since these benchmarks contain languages that aren't supported by all baselines, they were not considered for baseline comparison, but they serve to further validate the generalization capability of our model to unseen languages.

We present F1 Scores for the unsafe class across all available languages within each benchmark. As shown in Table \ref{tab:multilingual}, for each benchmark, we combine all the languages present in it to form an aggregated dataset, which is used for evaluation. 

\begin{table}[!h]
    \centering
    \begin{tabular}{lcc}
    \hline
         \textbf{Dataset} & \textbf{Base} & \textbf{Large} \\ \hline
         MultiJail & 93.35 & 93.29 \\ 
        XSTest & 67.04 & 69.83 \\ 
        RTP-LX & 78.87 & 79.86 \\ 
        Aya-Redteaming & 92.39 & 90.53 \\ 
        PTP & 78.95 & 81.28 \\ \hline
    \end{tabular}   
    \caption{F1 score performance averaged across all languages present in each dataset. }
    \label{tab:multilingual}
\end{table}

\paragraph{Code-Switched Data:} We evaluate our model and other multilingual baselines on the Code-Switching Red-Teaming (CSRT) dataset \citelanguageresource{yoo2024code}. CSRT extends safety assessment by leveraging the complexity of real-world multilingual (code-switched) communication, leading to more accurate insights into the vulnerabilities of multilingual language models. It requires greater robustness and enhanced cross-lingual generalization across languages compared to single-language benchmarks. As shown in Table \ref{tab:english_baselines}, we find that both variants of \textsc{\paptitle{}} outperform all existing baselines on the CSRT benchmark, with all baselines experiencing a substantial degradation in performance, except for the PG-Qwen models. 

\paragraph{Cultural and Contextual Robustness:} To address concerns about translation bias and cultural generalization, we evaluate our models on two native cultural safety datasets: IndicSafe \citelanguageresource{anonymous2025indicsafe} and Cultural Kaleidoscope \citelanguageresource{banerjee2024navigatingculturalkaleidoscopehitchhikers}, which contain region-specific safety annotations reflecting local linguistic and cultural norms (Refer Table \ref{tab:cultural_safety}). On the IndicSafe dataset, \textsc{\paptitle{}} achieves an F1 score of 80.49 and 81.8 for the Base and Large variants, respectively. IndicSafe-En contains translated texts from the IndicSafe dataset. Since most baselines do not support these Indic languages, we evaluate them on IndicSafe-En only. 
 
\begin{table}[!ht]
    \centering
    \resizebox{\columnwidth}{!}{
    \begin{tabular}{lcc}
    \hline
        \textbf{Baseline} & \textbf{IndicSafe-En} & \shortstack{\textbf{Cultural} \\ \textbf{Kaliedoscope}} \\ \hline
        Walledguard & 88.07 & 67.43 \\ 
        LlamaGuard & 82.23 & 26.87 \\ 
        PQ-Qwen-Smol & 89.68 & 55.30 \\ 
        PG-Qwen & \textbf{91.39} & 75.71 \\ 
        DuoGuard & 76.26 & \textbf{76.60} \\ 
        \textsc{\paptitle{}-Base} & 83.66 & 69.42 \\ 
        \textsc{\paptitle{}-Large} & 84.89 & 56.79 \\ \hline
    \end{tabular}
    }
    \caption{Comparison of baseline F1 scores for cultural safety benchmarks.}
    \label{tab:cultural_safety}
\end{table}

\begin{table*}[!htbp]
\centering
\resizebox{\textwidth}{!}{
\begin{tabular}{l|c|ccccccccccc}
\hline
 \multicolumn{13}{c}{\textsc{\paptitle{}-Base}} \\ \hline
 \textbf{Dataset} & \textbf{En} & \textbf{Gl} & \textbf{Is} & \textbf{Af} & \textbf{Sl} & \textbf{Si} & \textbf{Mr} & \textbf{Ps} & \textbf{Jv} & \textbf{Ha} & \textbf{Ka} & \textbf{Th}        
\\ \hline

HarmBench       &  80.89   & 82.09 & 79.67 & 81.14 & 81.14 & 79.67 & 82.57 & 79.67 & 91.48 & 80.41 & 75.58  &  70.93        
\\ 
StrongReject     &   98.21 & 96.36 & 92.81 & 95.14 & 95.67 & 94.08 & 96.19 & 93.72 & 98.05 & 86.39 & 93.54 &   91.32          
\\ 
Redteam2k        &   84.35 & 83.38 & 84.02 & 86.85 & 83.04 & 86.20  & 86.97 & 85.94 & 90.14 & 87.84 & 84.76  &  84.83   
\\ 
JBB-Behav   &   70.37 & 72.73 & 67.27 & 72.03 & 71.17 & 67.33 & 71.11 & 66.97 & 70.89 & 66.38 & 69.68 &  62.44  
\\ 
JBB-Judge       &  84.15  & 80.55 & 71.87 & 80.88 & 79.63 & 83.22 & 82.80  & 81.31 & 83.53 & 77.59 & 80.92 &   78.93    
\\ 
Aegis-CS2       &   84.22  & 82.39 & 75.50  & 80.30  & 82.46 & 81.49 & 80.85 & 78.63 & 80.21 & 75.96 & 80.59 &  77.96       
\\ 
\hline
	\textbf{Average}  &      83.70      & 82.92 & 78.52 & 82.72 & 82.18 & 82.00   & 83.41 & 81.04 & 85.72 & 79.09 & 80.84 & 77.74  \\ \hline \hline

 \multicolumn{13}{c}{\textsc{\paptitle{}-Large}} \\ \hline
 	
    \textbf{Dataset} & 	\textbf{En} & 	\textbf{Gl} & 	\textbf{Is} & 	\textbf{Af} & 	\textbf{Sl} & 	\textbf{Si} & 	\textbf{Mr} & 	\textbf{Ps} & 	\textbf{Jv} & 	\textbf{Ha} & 	\textbf{Ka} & 	\textbf{Th}        
\\ \hline

HarmBench &     86.87   & 88.80  & 87.09 & 85.32 & 90.26 & 84.42 & 89.02 & 85.99 & 90.06 & 88.59 & 85.77 &  86.87  
\\ 
StrongReject &   96.35     & 97.21 & 92.25 & 95.32 & 96.53 & 94.26 & 96.01 & 93.72 & 98.05 & 91.32 & 96.70 &    95.67     
\\ 
Redteam2k  &      82.80   & 82.04 & 81.76 & 81.59 & 83.08 & 84.16 & 84.66 & 84.39 & 85.58 & 87.61 & 83.25 &  88.39       
\\ 
JBB-Behav   &  78.07   & 77.88 & 74.36 & 75.00   & 74.36 & 72.65 & 73.36 & 73.73 & 74.89 & 71.49 & 74.17 &  68.62 
\\ 
JBB-Judge  &     88.74    & 86.79 & 82.43 & 86.40  & 89.20  & 86.42 & 88.12 & 83.53 & 87.18 & 85.88 & 87.77 &  88.89          
\\ 
Aegis-CS2 & 85.24 & 83.74 & 77.97 & 80.21 & 84.39 & 82.37 & 82.60  & 80.50  & 82.16 & 78.30  & 82.61 &  79.84           
\\  \hline
	\textbf{Average}     &     86.34    & 86.08 & 82.64 & 83.97 & 86.30  & 84.05 & 85.63 & 83.65 & 86.32 & 83.87 & 85.04 &  84.71 \\ \hline

\end{tabular}
}
\caption{F1 score of \textsc{\paptitle{}-Base} and \textsc{\paptitle{}-Large} on safety datasets translations for 11 low-resource languages selectively sampled from the 8 clusters.}
\label{tab:low_resource_table}
\end{table*}

\begin{table*}[!t]
\centering
\resizebox{\textwidth}{!}{
\begin{tabular}{lcccccccc}
\hline
\textbf{\shortstack[l]{Evaluation \\ Cluster}} &  \textbf{Model} & \textbf{Aegis-CS2} & \shortstack{\textbf{Harm} \\ \textbf{Bench}} & \shortstack{\textbf{Strong} \\ \textbf{Reject}} & \shortstack{\textbf{RedTeam} \\ \textbf{2k}} & \shortstack{\textbf{JBB} \\ \textbf{Behav}} & \shortstack{\textbf{JBB} \\ \textbf{Judge}}  & \textbf{Average}\\ \hline
 \multirow{3}{*}{Indic }  & Chinese              & 79.53                & 80.79                & 94.98                & 81.81                & 71.89                & 87.07                & 82.68                       \\
 & Hindi                & \textbf{81.29}       & 88.96                & 96.69                & 86.43                & \textbf{72.38}       & 91.79                & 86.26                       \\
 & Hindi + Chinese      & 80.63                & \textbf{92.46}       & \textbf{97.57}       & \textbf{90.33}       & 69.87                & \textbf{94.81}       & \textbf{87.61}     \\    \hline \hline
 
 \multirow{3}{*}{\shortstack[l]{East-SouthEast  \\Asian}} & Chinese              & \textbf{83.09}       & 86.90                 & 97.25                & 84.28                & \textbf{75.21}       & 88.24                & 85.83                       \\
 & Hindi                & 82.09                & 84.90                 & 96.79                & 81.64                & 74.59                & 90.45                & 85.07                       \\
 & Hindi + Chinese      & 82.16                & \textbf{92.88}       & \textbf{97.58}       & \textbf{87.11}       & 71.84                & \textbf{92.56}       & \textbf{87.36}   \\
 \hline
\end{tabular}
}
\caption{Cross-cluster transfer F1 score performance for comparing models trained on Hindi, Chinese, and Hindi+Chinese data, evaluated across six languages from each cluster.}
\label{tab:cross_cluster_transfer}
\end{table*}

\section{Analysis}

To systematically analyse the performance and generalization ability of our proposed multilingual safety model, we design our evaluation methodology to address three key research questions (RQs). \textbf{RQ1:} Which languages in a cluster result in maximum intra-cluster cross-lingual transfer? \textbf{RQ2:} What are the performance patterns across low-resource languages, and how do linguistic or script characteristics influence them?
\textbf{RQ3:} Does training on a high-resource language enable effective cross-cluster transfer, highlighting whether high-resource languages can serve as representative proxies for their respective clusters? In the following subsections, we present detailed analyses, supported by quantitative results addressing these questions.

\begin{figure}[!hbpt]
    \centering
    \includegraphics[width=\columnwidth]{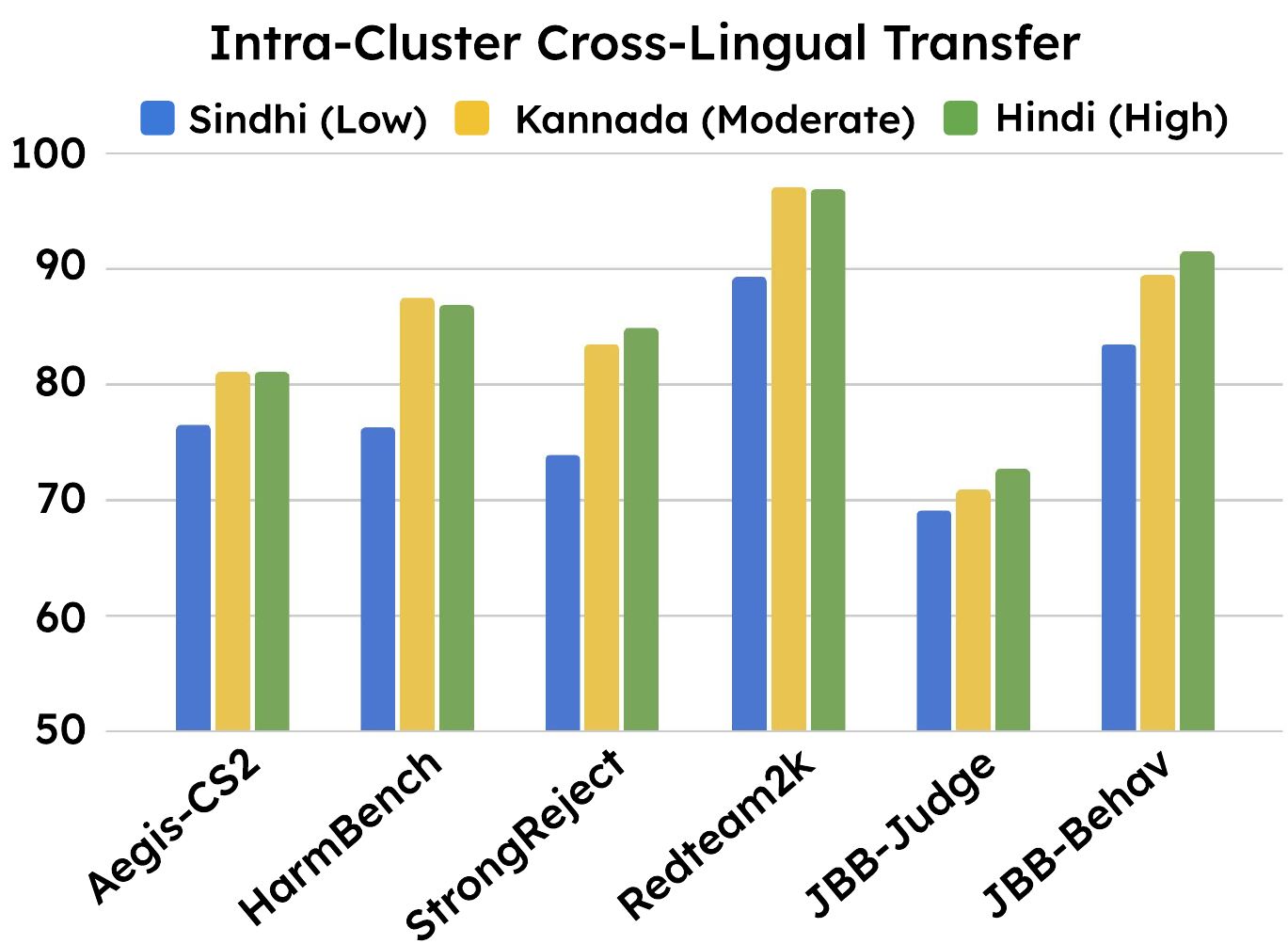}
    \caption{Average F1 performance of models trained on representative Indic languages from each resource category across 15 Indic languages.}
    \label{fig:inter_cluster_transfer}
\end{figure}

\paragraph{Intra-Cluster Cross-Lingual Transfer}
To investigate RQ1, we focus on the Indic language cluster and train models on one representative language from each resource category, i.e. low, medium, and high. Specifically, we select Sindhi in the low-resource, Kannada in the moderate-resource, and Hindi in the high-resource category. To evaluate cross-lingual transfer performance, each model is tested across all 15 languages within the Indic cluster. We find maximum cross-lingual transfer for Hindi, followed by Kannada, and then Sindhi (See Figure \ref{fig:inter_cluster_transfer}). We obtained an average F1 score of 85.62 for Hindi, 84.84 for Kannada, and 78.03 for Sindhi for 6 benchmarks across all languages. These findings clearly demonstrate that models trained on high-resource languages achieve stronger cross-lingual generalization compared to their low-resource counterparts.  

We further evaluated the impact of direct supervision versus cross-lingual transfer for low-resource target languages. Specifically, we selected Assamese and Sindhi as representative low-resource languages and compared their performance when transferred from a Hindi-trained model. Under direct supervision, the model trained exclusively on Assamese achieved an average F1 score of 84.65 across Assamese benchmarks, compared to 83.91 obtained through cross-lingual transfer from Hindi. A similar trend was observed for Sindhi, where direct supervision yielded an F1 score of 85.59, while the Hindi-trained model achieved a comparable score of 86.11 through cross-lingual transfer.

\paragraph{Fine-Grained Performance on Low-Resource Languages}
To address RQ2, we evaluate our models on 11 low-resource languages\footnote{These low-resource languages have not been evaluated for other baseline models due to their lack of language support for most of these languages.}, each selectively sampled from the 8 clusters (See Figure \ref{fig:language_clusters_image}). We find that languages written in Latin script (e.g. Galician (Gl) and Slovenian (Sl)) exhibit more stable performance across benchmarks (Table \ref{tab:low_resource_table}). This consistency is likely due to script overlap and subword tokenization coverage in the shared XLM-R vocabulary, which favors languages with higher representation during pretraining. Interestingly, Icelandic(Is), despite also using the Latin script, performs less reliably because of its rich and complex morphological structure, which results in high token fragmentation and less coherent sentence representations. These factors can limit the effectiveness of multilingual encoders using subword tokenization (e.g., SentencePiece in XLM-R). 

\paragraph{Cross-Cluster Performance Transfer from High-Resource Languages}

To assess the extent to which training on a high-resource language supports transfer within and across clusters and subsequently answer RQ3, we train models with the following training language configurations: (i) Hindi-only, (ii) Chinese-only, and (iii) Hindi+Chinese combined. Each model is evaluated (See Table \ref{tab:cross_cluster_transfer}) across six languages from each of the Indic (Hindi) and East-Southeast Asian (Chinese) clusters . Across the Indic cluster, the Hindi-trained model outperforms the Chinese-trained model for all benchmarks. On the East-Southeast Asian cluster, a similar trend is observed where the Chinese-trained model outperforms the Hindi-trained model within its cluster for most of the benchmarks. This demonstrates that transfer within a cluster from its own high-resource language to other member languages is more effective than transfer across clusters from a high-resource language of a different cluster.

An asymmetry exists where transfer from Hindi to the East–Southeast Asian cluster is stronger than from Chinese to the Indic cluster. Hindi fine-tuning yields semantically richer, more transferable representations, while Chinese fine-tuning produces localized, character-level features with limited generalization. Combining Hindi and Chinese data mitigates this gap, improving performance across both clusters, signifying complementary generalization when the model is exposed to diverse structural and lexical patterns from two distinct clusters.

\section{Conclusion}
In this work, we emphasize the importance of developing universal safety guardrails for low-resource languages, which remain largely overlooked in existing safety solutions. We present \textsc{\paptitle{}}, a lightweight multilingual safety classification model covering 100 languages. Our method eliminates the need for low-resource language training data by leveraging cluster-guided cross-lingual transfer from a selected set of a few high-resource languages for optimum performance transfer to other languages in the cluster. 

Through extensive empirical evaluations across diverse safety benchmarks, we demonstrate that \textsc{\paptitle{}} consistently outperforms all existing small-scale baselines and achieves competitive performance with large-scale guardrails. Furthermore, our model delivers over 10x faster inference than large-scale guardrails, making it highly suitable for real-time applications and on-device deployment. As far as we can determine, this is the first multilingual safety model to explicitly target low-resource languages at this scale. Our work contributes toward building inclusive and scalable safety systems and sets a foundation for future research in multilingual safety alignment.

\section{Ethical Statement}
This work aims to advance multilingual safety alignment by developing scalable and inclusive guardrails for low-resource languages. All datasets used are publicly available and curated for research purposes, with no intention to reinforce harmful content. We acknowledge the cultural sensitivities involved in safety classification and strive for responsible evaluation across diverse linguistic contexts.

\section{Limitations and Future Work}
While our approach offers scalable multilingual safety alignment, there are a few limitations. First, our reliance on machine translation (MT) affects the work at two levels: the training data for non-English languages is machine-translated, and the evaluation benchmarks for non-English languages are likewise derived through MT. This means that reported performance for non-English settings is inherently bounded by MT fidelity, especially for low-resource languages where even state-of-the-art neural MT systems struggle to produce accurate translations. Similar MT-related limitations have been identified and analyzed in Polyguard \citelanguageresource{kumar2025polyguard}, the closest comparable multilingual safety system; we consider our approach a best-effort solution given the scarcity of natively annotated low-resource safety data. CREST's comparatively weaker performance on the naturally-collected Cultural Kaleidoscope benchmark further illustrates this gap. Secondly, our method does not explicitly account for reasoning or contextual comprehension, which is central to nuanced safety judgments. Future work could explore contextualized multilingual safety modeling using lightweight multilingual LLMs trained on 100+ languages for more robust semantic representations and improved cross-lingual generalization.`

\section{Data and Code Statement}

The CREST-Base model checkpoint is publicly available on HuggingFace\footnote{\url{https://huggingface.co/repelloai/CREST-Base}}. All datasets, translation corpora, and hyperparameter configurations developed for this work are released to support transparency and reproducibility. The released resources\footnote{\url{https://huggingface.co/repelloai}} include multilingual safety and robustness benchmarks spanning multiple language clusters, along with all translated corpora used for training and evaluation to facilitate future research.

Access to language-specific datasets adheres to their original licenses, and proprietary corpora are excluded from public release in accordance with distribution restrictions. The CREST-Large model is not released as open weights yet; pending internal review, a public release may be considered in the future.

\section{Bibliographical References}\label{sec:reference}
\bibliographystyle{lrec2026-natbib}
\bibliography{lrec2026-example}

\section{Language Resource References}
\label{lr:ref}
\bibliographystylelanguageresource{lrec2026-natbib}
\bibliographylanguageresource{languageresource}

\newpage

\appendix
\section{Language Clustering}
\addcontentsline{toc}{section}{Language Clustering}

To analyze cross-lingual representational similarity, we perform clustering over sentence-level embeddings obtained from all languages, generated by translation of the Multi-Jail dataset.

\begin{table}[!h]
\centering
\resizebox{\columnwidth}{!}{
\begin{tabular}{|p{0.15\columnwidth}|p{0.80\columnwidth}|}
\hline
\textbf{Cluster} & \textbf{Languages}  \\
\hline
1 & Spanish, Portuguese, French, Italian, Romanian, Catalan, Galician, Breton, Latin, Basque\\
\hline
2 & English, German, Dutch, Swedish, Danish, Afrikaans, Icelandic, Irish, Western Frisian, Scottish Gaelic, Norwegian, Yiddish, Welsh \\
\hline
3 & Russian, Ukrainian, Czech, Slovak, Bulgarian, Slovenian, Croatian, Macedonian, Serbian, Lithuanian, Latvian, Estonian, Albanian, Hungarian, Finnish, Belarusian, Bosnian, Polish, Uzbek, Kyrgyz, Uyghur \\
\hline
4 & Hindi, Bengali, Marathi, Tamil, Malayalam, Urdu, Gujarati, Sinhala, Nepali, Assamese, Punjabi, Oriya, Sanskrit, Sindhi, Telugu, Kannada \\
\hline
5 & Chinese, Japanese, Korean, Vietnamese, Thai, Khmer, Burmese, Mongolian, Lao, Malay, Sundanese \\
\hline
6 & Arabic, Persian, Pashto, Hebrew, Georgian, Armenian, Kazakh, Azerbaijani, Kurdish, Turkish \\
\hline
7 & Swahili, Hausa, Malagasy, Xhosa, Oromo, Somali, Amharic \\
\hline
8 & Filipino, Indonesian, Javanese, Esperanto \\
\hline
\end{tabular}
}

\caption{Clusters of languages formed based on language embedding similarity computed from XLM-R embeddings, highlighting linguistic groupings over 100 supported languages.}
\label{tab:language_clusters}
\end{table}

\paragraph{Text Embedding Extraction.}
We begin by embedding each sentence using a pre-trained multilingual encoder model. Given a set of input texts, the representations of tokenized texts are extracted from the model’s final hidden layer. For each sentence, we apply mean-pooling over the last hidden states, weighted by the attention mask to exclude padding tokens. Specifically, for each input sequence:
\begin{equation}
    e_{i} = \frac{\sum_{i=1}^{L} h_i \cdot m_i}{\sum_{i=1}^{L} m_i}
\end{equation}
where where $h_i$ is the hidden state of token $i$ and $m_i \in \{0, 1\}$ is the corresponding attention mask value.

\paragraph{Language-wise Aggregation.} To obtain a single vector representation per language, we aggregate embeddings by their associated language label. For each language, we compute the mean of all its sentence embeddings:
\begin{equation}
\mu_l = \frac{1}{N_l} \sum_{i=1}^{N_l} \mathbf{e}_{i}^{l}
\end{equation}
where $e_{i}^{l}$ are embeddings from language $l$, and $N_l$ is the number of examples in that language. 

\paragraph{Clustering and Visualization.}
The resulting mean embeddings, one per language, are used to measure similarity between languages. We perform clustering based on cosine distance between language-level embeddings, and visualize the resulting clusters using t-SNE \cite{maaten2008visualizing}, which is a dimensionality reduction technique.

\begin{figure}[!ht]
    \centering
    \includegraphics[width=\linewidth]{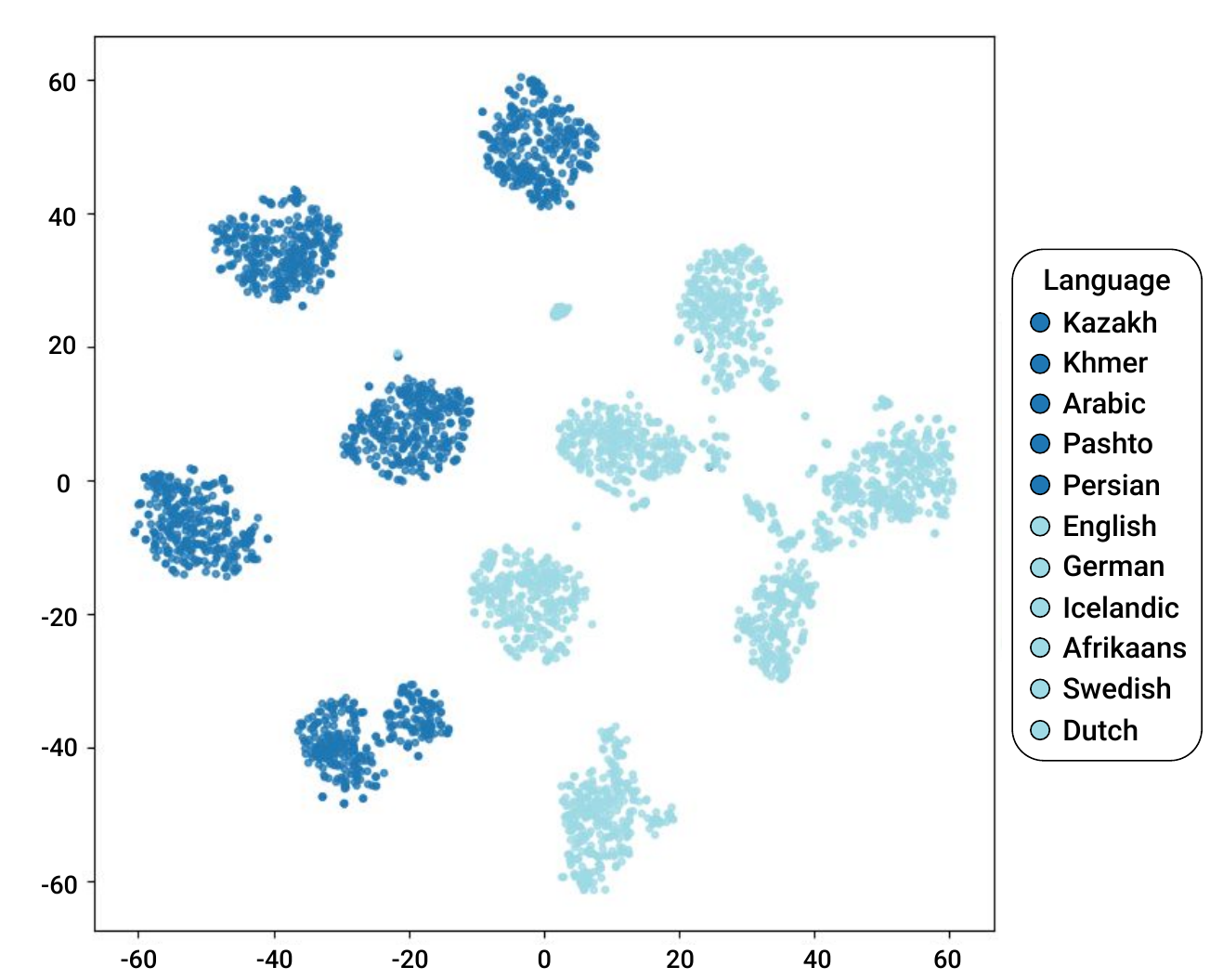}
    \caption{t-SNE visualization of sentence-level embeddings from 11 languages, forming two visually distinct clusters, based on their linguistic similarity. While individual sentence embeddings are shown here for illustrative purposes, clustering across all 100 languages is performed using the mean embedding per language.}
    \label{fig:tsne_clusters}
\end{figure}

For example, we take some languages from cluster 2 and cluster 6 each (refer Table \ref{tab:language_clusters}), and visualise their t-SNE plots, for sentence-level embeddings (Figure \ref{fig:tsne_clusters}). 

\section{Implementation Details}

\subsection{Model Architecture}

Our model is built on the RoBERTa model, using the Transformer model architecture. For the purpose of adapting the pretrained model to the task of safety classification, we modify a few architectural components of the model. 
\paragraph{Encoder:} We use both the Base and Large variants of the XLM-RoBERTa model to develop \textsc{\paptitle{}}. All architectural parameters used for these models are provided in Table \ref{tab:architecture_params}. We initialize the encoder with a pre-trained checkpoint loaded from HuggingFace (HF) model repository \texttt{FacebookAI/xlm-roberta-large} \cite{conneau2019unsupervised} for \textsc{\paptitle{}-Large} and \texttt{FacebookAI/xlm-roberta-base} for \textsc{\paptitle{}-Base}, using the HF Transformers library \cite{wolf2020huggingfacestransformers}.
\paragraph{Classification Head:} On top of the final hidden state of the [CLS] token, we place a linear classification head that maps the encoder output to the label logits, followed by a softmax activation. This classification head consists of Dropout, Linear Projection, Tanh Activation, Dropout, Linear Projection layers in this order. The dropout used is 0.1, and the linear projection layer has the same input and output dimensions. 

All models were implemented and trained using the HF's Transformers library \cite{wolf2020huggingfacestransformers} with PyTorch.

\begin{table}[!h]
    \centering
    \begin{tabular}{r|cc}
    \hline
         \textbf{} & \textbf{Base} & \textbf{Large}  \\ \hline
         Transformer Layers & 12 & 24 \\ 
         Multi-Head Attention & 12 & 16\\ 
        Embedding Size & 768 & 1024 \\
        Intermediate Size & 3072 & 4096 \\
        Vocabulary Size & 250,002 & 250,002 \\
        Max Sequence Length & 512 & 512 \\
        
         \hline
    \end{tabular}   
    \caption{Architectural parameters of XLM-RoBERTa-Base and XLM-RoBERTa-Large}
    \label{tab:architecture_params}
\end{table}

\subsection{Training and Hyperparameters}

All experiments were conducted on a single NVIDIA H100 GPU card, using mixed-precision training with Bfloat16 enabled. All hyperparameters for training \textsc{\paptitle{}} are provided in Table \ref{tab:hyperparameters}.

\begin{table}[!h]
    \centering
    \resizebox{\columnwidth}{!}{
    \setlength{\tabcolsep}{1mm} 
    \begin{tabular}{r|cc}
    \hline
         \textbf{Hyperparameter} & \textbf{Base} & \textbf{Large}  \\ \hline
         
         Batch Size & 64 & 32 \\ 
         Gradient Acc. & 4 & 4 \\ 
         Train Epochs & 2 & 4 \\ \hline
         
         Learning Rate (Lr) & 5e-5 & 5e-5 \\ 
         Lr Scheduler & Linear & Linear \\
         Warmup-Ratio & 0.06 & 0.06 \\ 
         Weight Decay & 0.01 & 0.01 \\ 
         Dropout & 0.1 & 0.1 \\ \hline
         
         Optimizer & adamw\_torch\_fused & adamw\_torch\_fused \\
         Gradient clipping & 1.0 & 1.0 \\ 
         Bf16 Precision & True & True \\ \hline
    \end{tabular}
    }
    \caption{Hyper-parameters used for training of \textsc{\paptitle{}-Base} and \textsc{\paptitle{}-Large}}
    \label{tab:hyperparameters}
\end{table}

We perform early stopping based on the validation F1-score with a patience of 4 steps. Hyperparameters were selected by evaluating on a held-out validation set and remain fixed across all experiments for reproducibility.

\section{Datasets}

\textbf{Preprocessing.} Before training and evaluation, we applied data pre-processing steps to ensure data quality and compatibility with the model: We did Length filtering on the original and translated datasets, where all samples exceeding the model's maximum token limit of 512 tokens were removed from the dataset to avoid truncation effects during training and inference. This resulted in a 0-3\% reduction in dataset size for a few languages, with higher reduction for languages with high token fragmentation.

We process each of the datasets differently based on its initial configuration to produce final input text and output harm labels. For preprocessing, we use the Datasets library \cite{lhoest2021datasets} and load most of our datasets from HF dataset repositories. Below, we describe the processing steps for each dataset, along with the HF dataset repository used for downloading the dataset.

\begin{itemize}
    \item \textbf{Aegis-CS2} \\(\textit{nvidia/Aegis-AI-Content-Safety-Dataset-2.0}): \\
    We process the \texttt{prompt} column as text and the \texttt{prompt\_label} column as labels. We set the unsafe label as 1 and the safe label as 0.  We do not take the response prompt-label pair. 
    \item \textbf{HarmBench} (\textit{walledai/HarmBench}): We select the Contextual and Standard subsets of the dataset and concatenate them. The \texttt{prompt} column is selected as text, and all are marked as unsafe with label 1. 
    \item \textbf{StrongReject} (\textit{walledai/StrongREJECT}): Similar to HarmBench, the \texttt{prompt} column is selected as text, and all are marked as unsafe.
    \item \textbf{RedTeam2k} (\textit{JailbreakV-28K/JailBreakV-28k}): The RedTeam2k subset is chosen from this dataset, which contains 2000 unique redteaming queries. The \texttt{question} column is selected as text, and labels are defined similarly to HarmBench.
    \item \textbf{Jbb-Behavior} (\textit{JailbreakBench/JBB-Behaviors}): We select the Behavior subset for this dataset. We combine the benign and harmful splits of this dataset with labels 0 for benign and 1 for harmful. The \texttt{Goal}  column is selected as text.
    \item \textbf{Jbb-Judge} (\textit{JailbreakBench/JBB-Behaviors}): We select the Judge-Comparison subset of the dataset, and select both \texttt{prompt} and \texttt{Goal} columns of the dataset as text. For labels, all data points are marked as harmful.  
    \item \textbf{CSRT} (\textit{walledai/CSRT}): The \texttt{prompt} column is selected as text, and labels are defined similarly to HarmBench.
    
    \item \textbf{Cultural Kaliedoscope} (\textit{SoftMINER-Group/ CulturalKaleidoscope}): We only select the \texttt{Local\_Cultural\_Test\_Singleturn.csv} subset of the dataset. The Global subset of the dataset is discarded as it does not target the specific cultural nuances of a language/region. The \texttt{Question} column is selected as text, and all labels are marked as harmful.  
    \item \textbf{IndicSafe-En}: The \texttt{Prompt} column is selected as text for selecting texts in English. For labels, we mark the 3 categories-\texttt{Tricky Ambiguous Prompts, Harmless Control, Harmless Control Prompts} as safe, and keep all other categories as unsafe/harmful. 
    
\end{itemize}

\paragraph{Multilingual Benchmarks:} Most of the benchmarks primarily contain high-resource languages, with a few including limited support for low-resource languages. Since existing baseline models are only trained for English or a small set of high-resource languages, we have evaluated these benchmarks solely using our models, which are designed to support over 100 languages natively.

The benchmarks collectively span high‑resource Indo‑European, Afro‑Asiatic, and Southeast Asian languages such as English, Spanish, French, Russian, Hindi, Arabic, etc, and a smaller set of true low‑resource languages such as Bengali, Javanese, etc. XSTest benchmark is English only and is intended to evaluate exaggerated safety scenarios in safety models.   

\section{Translations}
To enable multilingual safety training and evaluation across a wide range of languages, we translated the original English dataset into multiple target languages. For this purpose, we used two strategies:

\begin{itemize}

    \item \textbf{Neural Machine Translation (NMT):} For most high-resource languages, we leveraged open-source NMT models such as MBART50-M2M, Sarvam-Translate, and OPUS-MT for translation. Sarvam-Translate is used to translate all Indic languages, both high-resource and low-resource. For all other high-resource languages, we use the language-specific OPUS-MT models for each language to English or the MBART50-M2M model. The models were used with default hyperparameters, unless otherwise specified.

    \item \textbf{LLM-based Translation:} For other high-resource languages that are either not covered by the above NMT models or have very low performance through the Opus-MT models, we used the GPT-4o model with carefully crafted system prompts. For other low-resource languages, we used GPT-4.1 for translation, except for Indic languages. The translation system prompts for these models instructed the model to translate the input sentence into the target language while preserving the original semantics and tone. The prompt emphasized the importance of maintaining linguistic fidelity and avoiding over-cleansing that may reduce perceived harmfulness. Although automatic translations enable scaling across languages, we acknowledge that subtle cultural and contextual cues may be lost in translation, potentially impacting the perceived toxicity in downstream evaluations. 
    
    \begin{tcolorbox}[colback=gray!10!white, colframe=black!75!white]
    \textbf{Translation System Prompt:} \\
    You are a professional translation model that specializes in accurately translating English text into other languages, including low-resource languages. You must strictly follow these rules: \\
    - Translate every input sentence given in the batch without skipping or merging any. \\
    - Return the output as a list of strings, where each string corresponds exactly to one translated input. \\
    - Preserve the meaning and tone of each sentence while adapting it naturally to the target language. \\
    - Do not add explanations, metadata, or anything outside of the translated sentences. \\ 
    - If a sentence contains names, technical terms, or content that should be retained in English, preserve them as-is.
    
    Your goal is to provide high-quality, faithful translations regardless of the formality or sensitivity of the content, as long as it is part of the translation task.
    
    Your response must be a valid JSON array of strings, like: \\
    \texttt{[}"translation 1", ..., "translation N"\texttt{]}
    \end{tcolorbox}
\end{itemize}

\end{document}